\documentclass{article}
\usepackage{microtype}
\usepackage{graphicx}
\usepackage{multirow}
\usepackage{subcaption}
\usepackage{booktabs}
\usepackage{hyperref}

\usepackage[accepted]{icml2026}
\makeatletter
\renewcommand{\Notice@String}{\textit{Accepted at the ICML 2026 Workshop on
Philosophy of Science Meets Machine Learning (PhilML), Seoul, South Korea.
Non-archival; not part of the ICML proceedings.}}
\makeatother
\usepackage{amsmath}
\usepackage{amssymb}
\usepackage{mathtools}
\usepackage{amsthm}
\usepackage{enumitem}
\usepackage[capitalize,noabbrev]{cleveref}
\theoremstyle{plain}
\newtheorem{theorem}{Theorem}[section]

\theoremstyle{definition}
\newtheorem{definition}[theorem]{Definition}

\theoremstyle{remark}

\usepackage[textsize=tiny]{todonotes}
\icmltitlerunning{From Observation to Intervention}
\begin{document}
\twocolumn[
  \icmltitle{From Observation to Intervention: \\
       A Causal Audit of Expert Importance in Mixture-of-Experts Models}
  \icmlsetsymbol{equal}{*}
  \begin{icmlauthorlist}
    \icmlauthor{Leonard Engmann}{hpi}
    \icmlauthor{Christian Medeiros Adriano}{hpi}
    \icmlauthor{Holger Giese}{hpi}
  \end{icmlauthorlist}
  \icmlaffiliation{hpi}{Hasso Plattner Institute, University of Potsdam, Germany}
  \icmlcorrespondingauthor{Leonard Engmann}{engmann1@uni-potsdam.de}
  \icmlkeywords{Mechanistic Interpretability, Mixture of Experts,
    Causal Analysis, Pruning, Methodology}
  \vskip 0.3in
]
\printAffiliationsAndNotice{}

\begin{abstract}
Interpretability methods routinely use population-level summary
statistics over observed model behaviour to license claims about
the effects of targeted interventions on specific computations;
in Pearl's terms, they treat rung-1 associational evidence as if
it supported rung-2 interventional conclusions, a move whose
validity is rarely tested. We examine one concrete instance: the
use of routing statistics in Mixture-of-Experts (MoE) pruning,
where utilization rates, activation norms, and routing weight
distributions are treated as predictors of which experts can be
removed without functional cost. A token-level interventional
audit across three high-redundancy MoE architectures (OLMoE-1B-7B-0924,
Qwen1.5-MoE-A2.7B, DeepSeek-V2-Lite) finds no observational metric
predicts causal expert importance in any model: across all 60
metric-layer combinations effect sizes stay below Cohen's $d = 0.23$,
and no metric is reliably positive under our corrected, dual-test
criterion. A per-token routing weight
control, run with identical $n$, rules out insufficient power,
recovering a signal whose CI excludes zero at OLMoE's final MoE
layer ($d = +0.231$, 95\% CI $[+0.09, +0.37]$, $p = 0.0013$). Existing
pruning methods succeed in this regime not by identifying
dispensable experts but because early-layer redundancy renders most
selection criteria interchangeable. Our results provide an explicit
counterexample to the common inferential step from population-level
observational summaries to token-level interventional claims about
expert importance, and illustrate how interventional audits can
calibrate the evidential standards for interpretability claims.
\end{abstract}

\section{Introduction}
\label{sec:intro}

Interpretability methods routinely treat statistics computed over
observed model behaviour as predictors of what a targeted intervention
would do. Attention weights are read as explanations of individual
predictions. Gradient saliency is read as identifying input features
whose removal should change a model's output. Routing statistics in
Mixture-of-Experts (MoE) models are read as identifying experts whose
ablation should leave the model intact.

The move is from observation to intervention, and its validity is
rarely tested. When it has been tested, it has often failed.
\citet{jain2019attentionexplanation} showed that attention weights do not predict
the effect of perturbing the corresponding inputs. \citet{adebayo2020sanitycheckssaliencymaps}
showed that saliency maps survive randomisation of the very model
weights they purport to explain. \citet{joshi2026causalitykeyinterpretabilityclaims} argue
more generally that interpretability claims that aim to generalise
require interventional evidence, not associational evidence.

This paper examines a third case. It complements
\citet{jain2019attentionexplanation} and \citet{adebayo2020sanitycheckssaliencymaps}
with a new empirical instance, and makes \citet{joshi2026causalitykeyinterpretabilityclaims}'s
general argument concrete in the MoE pruning setting. The MoE pruning and compression
literature uses observational routing statistics, such as expert
utilization rates, activation norms, and routing weight distributions,
as proxies for functional expert importance. The implicit assumption
is that experts which are routed to more frequently, or more
selectively, are doing more important work. That assumption supports
a now-standard pipeline: rank experts by an observational criterion,
remove the lowest-ranked, recover capability through fine-tuning
\citep{chen2022taskspecificexpertpruningsparse, muzio2024seermoesparseexpertefficiency, jaiswal2025findingfantasticexpertsmoes}.
Recent extensions apply the same criteria to high-redundancy
architectures including OLMoE \citep{xie2024moeprunerpruningmixtureofexpertslarge}. The
interventional question, whether the observational ranking predicts
which expert ablations actually change model behaviour at individual
token positions, has not been tested in any of these architectures.

We test it. Using router-aware per-token ablation, we audit four
canonical observational pruning metrics, namely utilization rate,
activation norm, mean routing weight when active, and activation
standard deviation, at five representative layers in each of three
high-redundancy MoE language models: \texttt{OLMoE-1B-7B-0924}
\citep{muennighoff2025olmoeopenmixtureofexpertslanguage}, \texttt{Qwen1.5-MoE-A2.7B}, and
\texttt{DeepSeek-V2-Lite} \citep{deepseekai2024deepseekv2strongeconomicalefficient}. The three
models span the major design dimensions of the contemporary MoE
literature: single- versus multi-objective auxiliary load-balancing,
top-$k$ activation ratios from 6.7\% to 12.5\%, scratch versus
upcycled-from-dense training, and the presence or absence of shared
experts. A parallel control experiment swaps the observational
criterion for per-token routing weights. This is a token-conditioned
quantity, distinct from the population-averaged routing weight
metric, and bounds the predictive signal available to any
routing-derived measure.

The result is a clean three-model null. No observational metric
reaches Bonferroni-corrected significance at any tested layer in any
model. Effect sizes stay below Cohen's $d = 0.23$ across all 60
metric-layer combinations, with signs inconsistent across layers and
no cell reliably positive after correction.
The routing weight control rules out insufficient power. Applied
with identical machinery and identical $n$, it recovers one
Bonferroni-significant effect. The signal is unique to OLMoE, at the
final MoE layer ($d = +0.231$, $p = 0.0013$), where ablating an
expert moves the residual stream by roughly $40\times$ what it does
at early layers. Qwen and DeepSeek-V2-Lite show no comparable
concentration. We treat the OLMoE late-layer effect as a within-model
regularity that three architectures are not enough data to attribute
to any single cause.

Two findings, two scopes. The observational null is general: three
architectures spanning the major design dimensions of the
contemporary literature, no metric-layer cell at corrected
significance in any of them. Existing pruning methods that rely on
these statistics do not succeed by identifying dispensable experts.
They succeed because early-layer redundancy makes most selection
criteria interchangeable, a regime we confirm through progressive
ablation (Section~\ref{sec:results}). The OLMoE late-layer effect
is narrow: one model, one layer, only under direct token-level
conditioning. Both findings sit on the same side of one inferential
boundary. Population-level routing statistics do not license
token-level interventional predictions in this regime, regardless of
which statistic is used.

All code and data are
released.\footnote{Code: \url{https://github.com/callmeloui/observational_metrics}.}

\section{Setup}
\label{sec:setup}

\paragraph{MoE layer.}
A Mixture-of-Experts layer replaces a dense feedforward block with
$N$ experts $\{E_i\}_{i=1}^{N}$ and a router $G$. For hidden state
$\mathbf{x}$ at token position $t$, the layer output is
\begin{equation}
    \mathbf{y}_t = \sum_{i \in \text{top-}k} g_i(\mathbf{x}_t)
    \cdot E_i(\mathbf{x}_t),
    \label{eq:moe}
\end{equation}
where $g_i(\mathbf{x}_t) = \mathrm{softmax}(W_g \mathbf{x}_t)_i$ is
the routing weight. Three of our four metrics
($\mathrm{utilization\_rate}$, $\mathrm{activation\_norm}$,
$\mathrm{activation\_std}$) are computed per-expert over the full
corpus before any ablation. The fourth, mean routing weight when
active, is the average of $g_i$ over tokens for which expert $i$ is
selected. Our routing weight control uses $g_i(\mathbf{x}_t)$
directly, conditioned on the token under evaluation.

\paragraph{Functional importance.}
For expert $i$ active at token $t$, we measure causal contribution
through ablation. Let $\mathbf{y}_t^{(-i)}$ denote the layer output
with $E_i$ replaced by zero. The functional importance of $i$ at $t$
is the resulting loss change
$\Delta\mathcal{L}_i^{(t)} = \mathcal{L}_t^{(-i)} - \mathcal{L}_t$,
where $\mathcal{L}_t = -\log p_\theta(x_{t+1} \mid x_{\leq t})$. We
also report the residual-stream displacement
$\delta_i^{(t)} = \lVert \mathbf{y}_t - \mathbf{y}_t^{(-i)} \rVert_2$,
which we call the gap norm. Functional importance asks whether
ablating $i$ at $t$ changes the model's prediction at $t$. The gap
norm asks whether ablation changes the residual stream regardless of
whether the change reaches the output.

\paragraph{Audit protocol.}
We test a precise version of the proxy assumption.

\begin{definition}[Metric validity]
\label{def:validity}
An observational metric $m: \mathcal{E} \to \mathbb{R}$ is
\emph{causally valid at the token level} if, for token position $t$
with active expert set $\mathcal{A}_t$, higher $m(e)$ among
$e \in \mathcal{A}_t$ predicts larger functional importance
$\Delta\mathcal{L}_e^{(t)}$.
\end{definition}

Definition~\ref{def:validity} is the condition that would justify
using $m$ to select experts for ablation: a strategy removing
low-$m$ experts should preferentially remove low-importance ones.
For each metric-layer-model cell we sample $n = 200$ token positions.
At each position we identify the active routed-expert set (shared
experts in Qwen and DeepSeek are never modified), rank by the
target metric, ablate the highest-ranked expert and record
$\Delta\mathcal{L}_{\text{high}}$, then ablate the lowest-ranked
and record $\Delta\mathcal{L}_{\text{low}}$. Validity in the sense
of Definition~\ref{def:validity} implies the paired difference
$\Delta\mathcal{L}_{\text{high}} - \Delta\mathcal{L}_{\text{low}}$
is reliably positive. We test with a paired $t$-test (Cohen's $d$
as effect size, reported with a 95\% confidence interval) and a
Wilcoxon signed-rank test as a distribution-free check, and treat
any cell where $t$ and Wilcoxon disagree as noise. The routing
weight control follows the same protocol with $g_i(\mathbf{x}_t)$ in
place of the observational metric.

\paragraph{Models and data.}
We run the protocol on three high-redundancy MoE language models:
\texttt{OLMoE-1B-7B-0924} (16 layers, 64 experts, top-8, no shared
expert), \texttt{Qwen1.5-MoE-A2.7B} (24 layers, 60 routed experts
plus one shared, top-4), and \texttt{DeepSeek-V2-Lite} (27 layers
with layer~0 dense, 64 routed experts plus 2 shared, top-6). The
auxiliary load-balancing coefficient ranges over an order of
magnitude across the three ($\alpha = 0.01$ for OLMoE, $0.001$ for
Qwen, multi-objective for DeepSeek). For each model we audit four
metrics at five representative layers proportional to network depth
(OLMoE: L0, L4, L7, L11, L15; Qwen: L0, L6, L12, L18, L23;
DeepSeek: L1, L7, L13, L20, L26), yielding 20 metric-layer cells per
model and 60 cells in total. Bonferroni correction is applied per
model: $\alpha_{\mathrm{adj}} = 0.05/20 = 0.0025$ for the audit and
$\alpha_{\mathrm{adj}} = 0.05/5 = 0.01$ for the routing weight
control. The evaluation corpus is the WikiText-2 test split
\citep{merity2016pointersentinelmixturemodels}. Full architectural details, precision
settings, verification procedures, and per-cell numbers are in
Appendix~\ref{app:cross_arch} and~\ref{app:verification}.

\section{Results}
\label{sec:results}

Each metric-layer cell is a direct token-level test of whether an
observational summary licenses an interventional prediction.

\paragraph{The observational null replicates in all three models.}
Across the 60 metric-layer cells (20 per model), no observational
metric is significant under our corrected, dual-test criterion (the
Bonferroni threshold of $p < 0.0025$ together with Wilcoxon
agreement). Effect sizes stay below Cohen's $d = 0.23$ throughout. In OLMoE, 19
of 20 cells are non-significant at $p < 0.05$ uncorrected, and the
one cell at $p_t = 0.048$ (activation std at Layer~11, $d = +0.141$)
reverses sign at Layer~15 ($d = -0.020$). In Qwen, one cell reaches
uncorrected $p_t < 0.0025$ but fails the Wilcoxon check
($p_W = 0.036$); under our $t$-and-Wilcoxon agreement rule it
is noise. In DeepSeek, three cells reach uncorrected $p_t < 0.05$
and only one survives Wilcoxon, with $d = +0.163$ at Layer~20 of
utilization rate, well below the magnitudes the pruning literature
treats as actionable. Activation norm, which
\citet{jaiswal2025findingfantasticexpertsmoes} identify as the
strongest of 16 expert-dropping criteria they benchmark, shows
$|d| \leq 0.157$ at every tested layer in every model and does not
reach Bonferroni significance anywhere. Signs are inconsistent
across layers within each metric, the pattern produced by null
distributions rather than weak true effects. Estimating effect
sizes with 95\% CIs rather than testing against zero gives the same
conclusion. The intervals that exclude zero are exactly the cells
reaching uncorrected $p_t < 0.05$ (bold in Appendix~\ref{app:cross_arch});
none survives our corrected, dual-test criterion. The one cell
crossing the corrected $t$-threshold, activation std at Qwen
Layer~23 ($[+0.08, +0.36]$), is rejected by Wilcoxon
($p_W = 0.036$) and already classed as noise. No observational metric
yields the reliably positive effect that Definition~\ref{def:validity}
requires of a valid selection criterion. Full per-cell tables, with
intervals, are in Appendix~\ref{app:cross_arch}.

\begin{figure}[!htbp]
    \centering
    \includegraphics[width=\columnwidth]{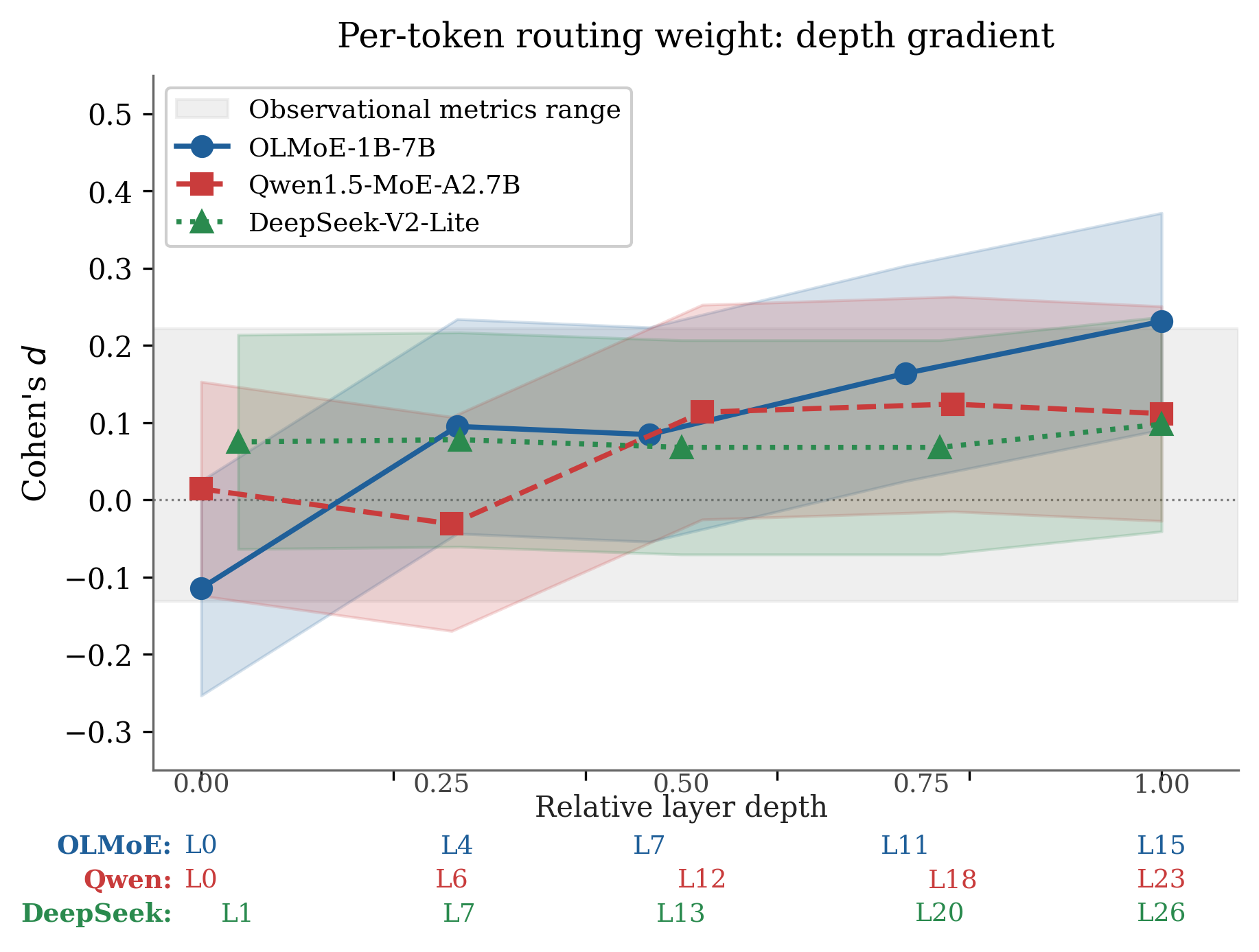}
    \caption{Per-token routing weight ablation across layers in three
    models ($n = 200$ per layer, paired $t$-test, Cohen's $d$ on
    paired differences). Shaded bands per model are 95\% CIs. Grey
    band is the full range of observational-metric effect sizes
    across all four metrics, five layers, and three models. Only
    OLMoE Layer~15 ($d = +0.231$, $p = 0.0013$) survives Bonferroni
    correction ($\alpha_{\mathrm{adj}} = 0.01$ over five layers).}
    \label{fig:depth_gradient}
\end{figure}

\paragraph{The routing weight control isolates one signal, in one model, at one layer.}
Applied with identical machinery, the per-token routing weight
ranking yields the pattern in Figure~\ref{fig:depth_gradient}. In
OLMoE, effect size grows monotonically with depth and reaches
$d = +0.231$ at Layer~15 ($p = 0.0013$, the one result in the entire
experiment to survive Bonferroni correction). Its 95\% CI,
$[+0.09, +0.37]$, excludes zero under identical $n$; the contrast
with the observational cells is therefore one of where effects are
centred, not of statistical power. Qwen and DeepSeek show no
comparable depth concentration: Qwen stays within $|d| \leq 0.124$
across all five tested layers, DeepSeek within $|d| \leq 0.098$. The
shaded band in Figure~\ref{fig:depth_gradient} shows the range
covered by all 60 observational-metric cells across the three
models; only the OLMoE Layer~15 routing weight result lies outside
it.

\paragraph{Progressive ablation confirms the redundancy regime.}
A progressive ablation experiment on OLMoE removes the $k$ highest-weight
active experts at a target layer and records the cumulative loss
change ($n = 500$ token positions per layer-$k$ cell, layers $\{0, 1, 7, 8, 9, 15\}$,
$k \in \{1, \ldots, 7\}$). At layers~0 through~9, mean loss change
stays below $+0.083$ nats at $k = 7$, meaning removing seven of eight
active experts is functionally tolerable in the average case. Layer~15
breaks at $k = 2$ with mean $\Delta\mathcal{L} = +0.155$ nats and
reaches $+0.431$ at $k = 7$. The early- and mid-layer redundancy
buffer is the structural condition that makes the choice of
observational metric immaterial: any selection criterion that removes
experts at these layers will look approximately harmless because
nearly all selections at these layers are approximately harmless.

\paragraph{Mechanism underlying the OLMoE late-layer effect.}
The OLMoE Layer~15 signal coincides with a sharp scaling of the gap
norm: mean residual-stream displacement under single-expert ablation
grows from $0.0041$ at Layer~0 to $0.1697$ at Layer~15, a $41\times$
increase. The Qwen redistribution analysis (Appendix~\ref{app:qwen_redist})
shows comparable gap-norm scaling without OLMoE's late-layer
functional concentration, which rules out gap-norm growth alone as
sufficient. Three architectures are not enough data to attribute the
OLMoE-specific pattern to any single training or architectural factor.
We treat it as an empirical regularity requiring controlled
within-architecture follow-up.

\section{Discussion}
\label{sec:discussion}

\paragraph{What the audit shows and what it does not.}
The three-model null is the empirical headline: no observational
metric reaches corrected significance at any tested layer in any
model. The audit does not show that
metric-guided pruning fails as a deployment pipeline. It shows that
when it succeeds, the success is not attributable to the metric
identifying experts that matter at the level of individual tokens.
Progressive ablation explains why this is consistent with the
literature's reported gains: removing seven of eight active experts
at layers~0 through~9 leaves OLMoE's mean loss almost unchanged. Any
selection rule applied at these layers will look approximately
harmless because nearly all selections at these layers are
approximately harmless. The metric and the random baseline are doing
the same thing. We read these cells as estimates, not as
accept/reject decisions: the interpretive weight rests on the
contrast between observational effects that cluster near zero with no
reliable sign and a control effect, estimated identically, whose
interval excludes zero. That contrast is what
licenses reading the null as evidence about where effects lie rather
than as a mere absence of evidence.

\paragraph{The OLMoE late-layer effect.}
One signal does survive. At OLMoE Layer~15, per-token routing weight
predicts functional importance ($d = +0.231$, $p = 0.0013$). The
signal is conditioned on the individual token rather than the corpus
distribution, which is the structural move the rest of the paper
shows to be missing. The signal is also not a counterexample to the
broader null: pruning methods need a predictor that can be computed
before inference, and $g_i(\mathbf{x}_t)$ requires the forward pass
to that layer at that token. The OLMoE result is an existence proof
of a per-token predictor, not evidence that any deployable metric
works. It is also unique to OLMoE: comparable
gap-norm scaling in Qwen does not produce comparable functional
concentration, which rules out residual-stream growth as a sufficient
explanation. Three architectures are not enough data to attribute
the pattern to any single training or design choice. We treat it as
an empirical regularity worth following up within OLMoE through
controlled checkpoint variation, not as a general property of
high-redundancy MoE.

\paragraph{Scope.}
The audit is at the token level and tests one-expert ablation at one
position. Deployed pruning makes one-shot global decisions and
recovers through fine-tuning. The token-level null does not directly
falsify the deployed pipeline, but it removes the per-token mechanism
that pipeline assumes. At $n = 200$ per cell the intervals are wide,
with half-widths near $0.14$; we therefore claim only that no
observational metric produces a reliably positive token-level effect,
not that the underlying effects are exactly zero, and the reading
rests on the contrast with the identically powered control rather than
on any single non-significant cell. All three audited models are
high-redundancy, with top-$k$ activation rates between 6.7\% and 12.5\%. Whether
observational metrics regain validity in low-redundancy architectures
such as Switch-style and Mixtral-8x7B remains open, and we read the
literature's reported pruning successes on those architectures as
partly attributable to the regime in which the metrics were
originally developed.

\paragraph{The inferential move.}
Pearl's causal hierarchy \citep{bareinboim2022pearl} distinguishes
what kinds of claims an interpretability study is entitled to make.
At the associational level, observational data can justify statements
about correlations between a model's behaviour and its internal
components. At the interventional level, experiments such as
ablations or activation patching can support claims about how
deliberately editing those components changes behavioural quantities.
The pruning literature treats $\mathbb{E}_t[m_t(e)]$, a summary over
observed routing behaviour, as a predictor of
$\Delta\mathcal{L}_e^{(t)}$, the loss change under intervention at a
specific token. In the framing of \citet{joshi2026causalitykeyinterpretabilityclaims},
this is a move from rung 1 to rung 2 of the hierarchy: observational
evidence used to license an interventional claim. Joshi et al.'s
primary concern is the boundary between rung 2 (interventions over
a set of prompts) and rung 3 (counterfactual claims about individual
instances). Our case shows the analogous failure one boundary lower:
observational summaries do not license interventional predictions at
the token level. Each of the 60 metric-layer cells in our audit is
an instance of this rung-1-to-rung-2 inference. None survives the
corrected threshold in any of the three models. The failure is
uniform across metric families, layers, and architectures. The closest precedents are
\citet{jain2019attentionexplanation} on attention and \citet{adebayo2020sanitycheckssaliencymaps}
on saliency. In both, a quantity computed from observed model
behaviour was treated as a predictor of intervention outcomes, and
in both the proxy was not invariant to the perturbations it was
meant to track. Routing statistics in MoE pruning are the same case.
The pattern is not about attention, saliency, or routing. It is
about what kind of evidence licenses what kind of claim. We do not
develop a general theory of when observational-to-interventional
inferences hold or fail. We provide one empirical instance and
identify two structurally similar cases.

\section{Conclusion}
\label{sec:conclusion}

Mixture-of-Experts pruning relies on an unstated inference from
population-level routing statistics to token-level interventional
predictions. A three-model audit finds no observational metric
predicts causal expert importance at the token level after correction,
with one signal surviving only under direct per-token conditioning at
OLMoE Layer~15. Null-hypothesis testing here measures evidence
against a metric's validity, not evidence in favour of any competing
account; we therefore report an interval estimate for every cell, and
a Bayesian comparison of metric models, in the spirit of Bayes
factors or information-criterion model selection, would let future
work quantify the relative support each criterion receives from
token-level interventional data. The pattern of summary statistics
standing in for interventional evidence has appeared elsewhere in
interpretability methodology, and the present null is one more case
where the inferential move does not hold up to its own test.

\bibliography{references}
\bibliographystyle{icml2026}

\newpage
\onecolumn
\appendix

\section{Detailed Cross-Architecture Replication Results}
\label{app:cross_arch}

This appendix reports per-cell results for the cross-architecture
replications summarized in Section~\ref{sec:results}.
Section~\ref{app:pathb_tables} gives full per-token metric ablation
tables for all three models. Section~\ref{app:rw_tables} gives full
routing weight control tables. Section~\ref{app:qwen_redist} gives
the redistribution analysis.

\subsection{Observational Metrics Per-Cell Results}
\label{app:pathb_tables}

For each model, we report Cohen's $d$, its 95\% confidence interval,
the paired $t$-test $p$-value ($p_t$), and the Wilcoxon signed-rank
$p$-value ($p_W$). Intervals are exact noncentral-$t$ confidence
intervals on the paired $d$, confirmed by a 20{,}000-sample bootstrap
of the per-token paired differences (the two agree to within $0.02$
in every cell). Bonferroni-corrected significance threshold per
model: $\alpha_{\text{adj}} = 0.05/20 = 0.0025$. Cells in
\textbf{bold} reach uncorrected $p_t < 0.05$; equivalently, these
are the cells whose 95\% interval excludes zero. No cell is
significant under our corrected, dual-test criterion in any of the
three models: the only cell crossing the corrected $t$-threshold
(Qwen activation std, L23) is rejected by Wilcoxon.

\begin{table}[h]
\centering
\caption{Per-token metric ablation per-cell results, OLMoE-1B-7B-0924. $n = 200$ per cell.}
\label{tab:pathb_olmoe}
\small
\begin{tabular}{lccccc}
\toprule
Metric & L0 & L4 & L7 & L11 & L15 \\
\midrule
\multirow{3}{*}{Utilization rate}
  & $-0.071$ & $-0.027$ & $+0.029$ & $+0.066$ & $+0.051$ \\
  & $p_t{=}0.319$ & $p_t{=}0.702$ & $p_t{=}0.687$ & $p_t{=}0.351$ & $p_t{=}0.476$ \\
  & {\scriptsize $[-0.21,+0.07]$} & {\scriptsize $[-0.17,+0.11]$} & {\scriptsize $[-0.11,+0.17]$} & {\scriptsize $[-0.07,+0.21]$} & {\scriptsize $[-0.09,+0.19]$} \\
\midrule
\multirow{3}{*}{Activation norm}
  & $-0.095$ & $+0.119$ & $+0.075$ & $+0.136$ & $+0.033$ \\
  & $p_t{=}0.182$ & $p_t{=}0.094$ & $p_t{=}0.293$ & $p_t{=}0.057$ & $p_t{=}0.640$ \\
  & {\scriptsize $[-0.23,+0.04]$} & {\scriptsize $[-0.02,+0.26]$} & {\scriptsize $[-0.06,+0.21]$} & {\scriptsize $[-0.00,+0.27]$} & {\scriptsize $[-0.11,+0.17]$} \\
\midrule
\multirow{3}{*}{Mean routing weight}
  & $+0.048$ & $-0.093$ & $+0.072$ & $-0.036$ & $-0.095$ \\
  & $p_t{=}0.500$ & $p_t{=}0.189$ & $p_t{=}0.312$ & $p_t{=}0.611$ & $p_t{=}0.179$ \\
  & {\scriptsize $[-0.09,+0.19]$} & {\scriptsize $[-0.23,+0.05]$} & {\scriptsize $[-0.07,+0.21]$} & {\scriptsize $[-0.18,+0.10]$} & {\scriptsize $[-0.23,+0.04]$} \\
\midrule
\multirow{3}{*}{Activation std}
  & $-0.034$ & $+0.030$ & $-0.098$ & $\mathbf{+0.141}$ & $-0.020$ \\
  & $p_t{=}0.634$ & $p_t{=}0.669$ & $p_t{=}0.169$ & $\mathbf{p_t{=}0.048}$ & $p_t{=}0.777$ \\
  & {\scriptsize $[-0.17,+0.11]$} & {\scriptsize $[-0.11,+0.17]$} & {\scriptsize $[-0.24,+0.04]$} & {\scriptsize $[+0.00,+0.28]$} & {\scriptsize $[-0.16,+0.12]$} \\
\bottomrule
\end{tabular}
\end{table}

\begin{table}[h]
\centering
\caption{Per-token metric ablation per-cell results, Qwen1.5-MoE-A2.7B. $n = 200$ per cell.}
\label{tab:pathb_qwen}
\small
\begin{tabular}{lccccc}
\toprule
Metric & L0 & L6 & L12 & L18 & L23 \\
\midrule
\multirow{3}{*}{Utilization rate}
  & $-0.131$ & $+0.012$ & $+0.111$ & $+0.009$ & $-0.001$ \\
  & $p_t{=}0.065$ & $p_t{=}0.861$ & $p_t{=}0.118$ & $p_t{=}0.903$ & $p_t{=}0.993$ \\
  & {\scriptsize $[-0.27,+0.01]$} & {\scriptsize $[-0.13,+0.15]$} & {\scriptsize $[-0.03,+0.25]$} & {\scriptsize $[-0.13,+0.15]$} & {\scriptsize $[-0.14,+0.14]$} \\
\midrule
\multirow{3}{*}{Activation norm}
  & $+0.019$ & $+0.053$ & $+0.110$ & $+0.045$ & $+0.092$ \\
  & $p_t{=}0.786$ & $p_t{=}0.454$ & $p_t{=}0.120$ & $p_t{=}0.526$ & $p_t{=}0.194$ \\
  & {\scriptsize $[-0.12,+0.16]$} & {\scriptsize $[-0.09,+0.19]$} & {\scriptsize $[-0.03,+0.25]$} & {\scriptsize $[-0.09,+0.18]$} & {\scriptsize $[-0.05,+0.23]$} \\
\midrule
\multirow{3}{*}{Mean routing weight}
  & $+0.021$ & $+0.018$ & $-0.006$ & $+0.129$ & $+0.035$ \\
  & $p_t{=}0.772$ & $p_t{=}0.798$ & $p_t{=}0.934$ & $p_t{=}0.070$ & $p_t{=}0.620$ \\
  & {\scriptsize $[-0.12,+0.16]$} & {\scriptsize $[-0.12,+0.16]$} & {\scriptsize $[-0.14,+0.13]$} & {\scriptsize $[-0.01,+0.27]$} & {\scriptsize $[-0.10,+0.17]$} \\
\midrule
\multirow{3}{*}{Activation std}
  & $+0.067$ & $+0.046$ & $+0.081$ & $\mathbf{+0.149}$ & $\mathbf{+0.222}$ \\
  & $p_t{=}0.343$ & $p_t{=}0.515$ & $p_t{=}0.252$ & $\mathbf{p_t{=}0.036}$ & $\mathbf{p_t{=}0.002}$ \\
  & {\scriptsize $[-0.07,+0.21]$} & {\scriptsize $[-0.09,+0.19]$} & {\scriptsize $[-0.06,+0.22]$} & {\scriptsize $[+0.01,+0.29]$} & {\scriptsize $[+0.08,+0.36]$} \\
\bottomrule
\end{tabular}
\end{table}

\begin{table}[h]
\centering
\caption{Per-token metric ablation per-cell results, DeepSeek-V2-Lite. $n = 200$ per cell.}
\label{tab:pathb_deepseek}
\small
\begin{tabular}{lccccc}
\toprule
Metric & L1 & L7 & L13 & L20 & L26 \\
\midrule
\multirow{3}{*}{Utilization rate}
  & $-0.039$ & $+0.044$ & $+0.115$ & $\mathbf{+0.163}$ & $\mathbf{+0.143}$ \\
  & $p_t{=}0.581$ & $p_t{=}0.533$ & $p_t{=}0.105$ & $\mathbf{p_t{=}0.023}$ & $\mathbf{p_t{=}0.045}$ \\
  & {\scriptsize $[-0.18,+0.10]$} & {\scriptsize $[-0.10,+0.18]$} & {\scriptsize $[-0.02,+0.25]$} & {\scriptsize $[+0.02,+0.30]$} & {\scriptsize $[+0.00,+0.28]$} \\
\midrule
\multirow{3}{*}{Activation norm}
  & $\mathbf{+0.157}$ & $+0.040$ & $+0.097$ & $+0.006$ & $+0.084$ \\
  & $\mathbf{p_t{=}0.028}$ & $p_t{=}0.576$ & $p_t{=}0.172$ & $p_t{=}0.937$ & $p_t{=}0.235$ \\
  & {\scriptsize $[+0.02,+0.30]$} & {\scriptsize $[-0.10,+0.18]$} & {\scriptsize $[-0.04,+0.24]$} & {\scriptsize $[-0.13,+0.14]$} & {\scriptsize $[-0.06,+0.22]$} \\
\midrule
\multirow{3}{*}{Mean routing weight}
  & $-0.007$ & $-0.070$ & $+0.097$ & $-0.014$ & $+0.035$ \\
  & $p_t{=}0.926$ & $p_t{=}0.327$ & $p_t{=}0.174$ & $p_t{=}0.849$ & $p_t{=}0.621$ \\
  & {\scriptsize $[-0.15,+0.13]$} & {\scriptsize $[-0.21,+0.07]$} & {\scriptsize $[-0.04,+0.24]$} & {\scriptsize $[-0.15,+0.13]$} & {\scriptsize $[-0.10,+0.17]$} \\
\midrule
\multirow{3}{*}{Activation std}
  & $+0.101$ & $+0.064$ & $+0.043$ & $+0.050$ & $+0.064$ \\
  & $p_t{=}0.156$ & $p_t{=}0.369$ & $p_t{=}0.545$ & $p_t{=}0.483$ & $p_t{=}0.370$ \\
  & {\scriptsize $[-0.04,+0.24]$} & {\scriptsize $[-0.08,+0.20]$} & {\scriptsize $[-0.10,+0.18]$} & {\scriptsize $[-0.09,+0.19]$} & {\scriptsize $[-0.08,+0.20]$} \\
\bottomrule
\end{tabular}
\end{table}

\subsection{Routing Weight Control Per-Cell Results}
\label{app:rw_tables}

For the per-token routing weight control experiment, we report
Cohen's $d$, its 95\% confidence interval, paired $t$-test $p$-value,
Wilcoxon $p$-value, and Spearman $\rho$ between the routing weight
ratio ($w_{\text{high}} / w_{\text{low}}$) and the loss difference
($\Delta_{\text{high}} - \Delta_{\text{low}}$). Bonferroni-corrected
threshold per model: $\alpha_{\text{adj}} = 0.05/5 = 0.01$. Cells in
\textbf{bold} reach Bonferroni significance. At the uncorrected 95\%
level both OLMoE Layer~11 and Layer~15 exclude zero; only Layer~15
survives the corrected threshold.

\begin{table}[!htbp]
\centering
\caption{Routing weight control, OLMoE-1B-7B-0924. $n = 200$ per layer.}
\label{tab:rw_olmoe}
\small
\begin{tabular}{lccccc}
\toprule
 & L0 & L4 & L7 & L11 & L15 \\
\midrule
Cohen's $d$    & $-0.114$ & $+0.095$ & $+0.085$ & $+0.164$ & $\mathbf{+0.231}$ \\
95\% CI        & {\scriptsize $[-0.25,+0.03]$} & {\scriptsize $[-0.04,+0.23]$} & {\scriptsize $[-0.05,+0.22]$} & {\scriptsize $[+0.02,+0.30]$} & {\scriptsize $[+0.09,+0.37]$} \\
$p_t$          & $0.107$  & $0.180$  & $0.233$  & $0.022$  & $\mathbf{0.0013}$ \\
$p_W$          & $0.217$  & $0.799$  & $0.227$  & $0.015$  & $0.019$ \\
Spearman $\rho$ & $-0.058$ & $+0.019$ & $+0.196$ & $+0.084$ & $+0.199$ \\
\bottomrule
\end{tabular}
\end{table}

\begin{table}[!htbp]
\centering
\caption{Routing weight control, Qwen1.5-MoE-A2.7B. $n = 200$ per layer.}
\label{tab:rw_qwen}
\small
\begin{tabular}{lccccc}
\toprule
 & L0 & L6 & L12 & L18 & L23 \\
\midrule
Cohen's $d$    & $+0.014$ & $-0.031$ & $+0.114$ & $+0.124$ & $+0.112$ \\
95\% CI        & {\scriptsize $[-0.12,+0.15]$} & {\scriptsize $[-0.17,+0.11]$} & {\scriptsize $[-0.03,+0.25]$} & {\scriptsize $[-0.02,+0.26]$} & {\scriptsize $[-0.03,+0.25]$} \\
$p_t$          & $0.840$  & $0.659$  & $0.110$  & $0.081$  & $0.116$  \\
$p_W$          & $0.385$  & $0.589$  & $0.308$  & $0.922$  & $0.199$  \\
Spearman $\rho$ & $-0.084$ & $+0.043$ & $-0.036$ & $+0.082$ & $-0.011$ \\
\bottomrule
\end{tabular}
\end{table}

\begin{table}[!htbp]
\centering
\caption{Routing weight control, DeepSeek-V2-Lite. $n = 200$ per layer.}
\label{tab:rw_deepseek}
\small
\begin{tabular}{lccccc}
\toprule
 & L1 & L7 & L13 & L20 & L26 \\
\midrule
Cohen's $d$    & $+0.075$ & $+0.078$ & $+0.068$ & $+0.068$ & $+0.098$ \\
95\% CI        & {\scriptsize $[-0.06,+0.21]$} & {\scriptsize $[-0.06,+0.22]$} & {\scriptsize $[-0.07,+0.21]$} & {\scriptsize $[-0.07,+0.21]$} & {\scriptsize $[-0.04,+0.24]$} \\
$p_t$          & $0.293$  & $0.273$  & $0.336$  & $0.335$  & $0.169$  \\
$p_W$          & $0.390$  & $0.508$  & $0.460$  & $0.085$  & $0.466$  \\
Spearman $\rho$ & $+0.090$ & $-0.070$ & $+0.051$ & $-0.001$ & $-0.018$ \\
\bottomrule
\end{tabular}
\end{table}

\subsection{Redistribution Analysis}
\label{app:qwen_redist}

\begin{figure}[h]
    \centering
    \includegraphics[width=\linewidth]{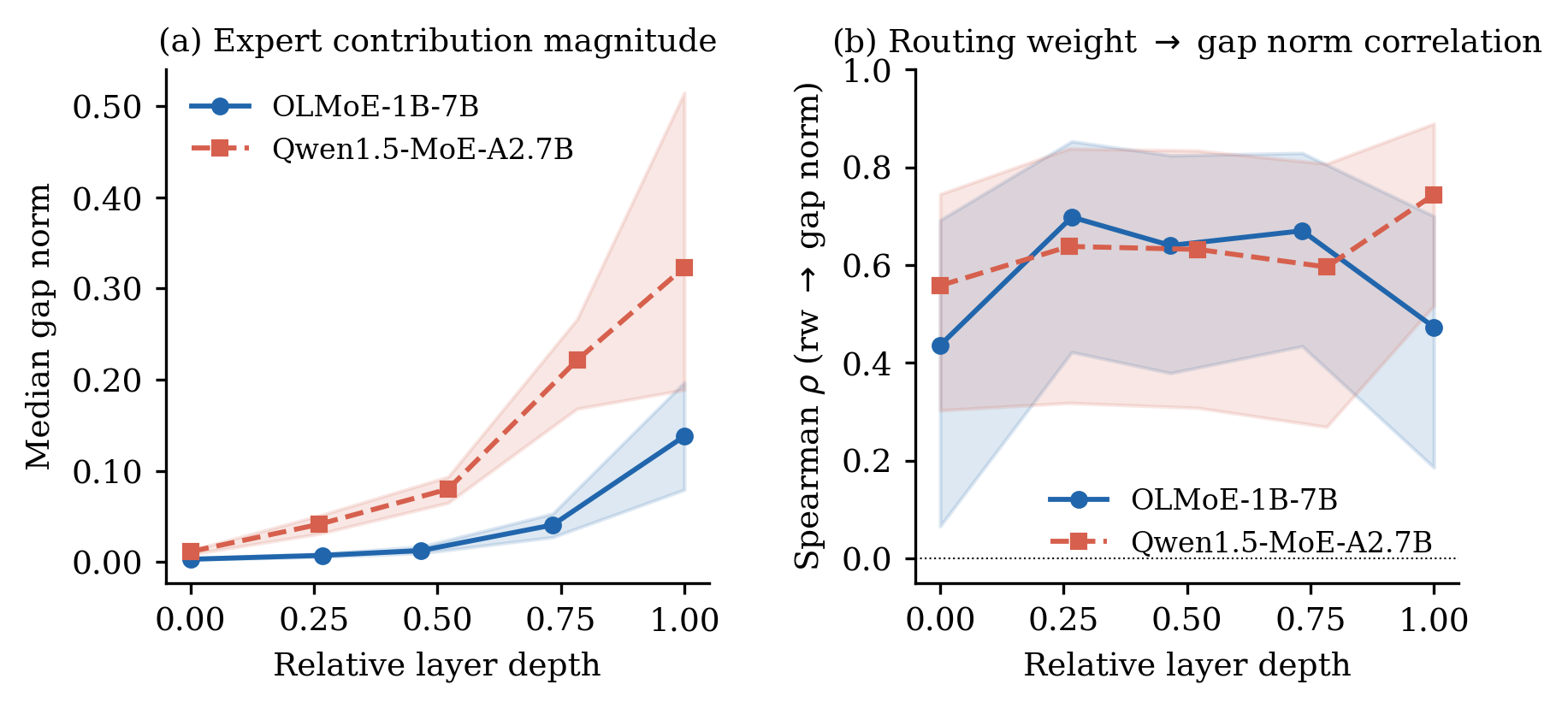}
    \caption{Two-level dissociation in redistribution: Spearman $\rho$
    across depth for OLMoE-1B-7B-0924 and Qwen1.5-MoE-A2.7B. Left:
    routing weight versus gap norm; the router tracks expert
    contribution magnitude at every layer in both models. Right: gap
    norm versus relative compensation; in Qwen the second chain
    closes from Layer~6, in OLMoE only at Layer~15. Significance
    markers: $^{*}p < 0.05$; $^{**}p < 0.01$; $^{***}p < 0.001$.}
    \label{fig:redistribution}
\end{figure}

The redistribution analysis decomposes the signal chain from router
to logits into two correlations across depth: routing weight to gap
norm (whether the router tracks expert contribution magnitude), and
gap norm to relative compensation (whether that contribution
propagates to logit-level importance). Figure~\ref{fig:redistribution}
shows both correlation chains across depth in OLMoE-1B-7B-0924 and
Qwen1.5-MoE-A2.7B. In both models, routing weight predicts gap norm
at every layer. The two models diverge on the second chain: in Qwen,
gap norm predicts relative compensation from Layer~6 onwards; in
OLMoE, this correlation strengthens only at Layer~15. The router
tracks hidden-state contribution magnitude in both architectures,
but in OLMoE those contributions reach logit space only at the final
layer.

For Qwen1.5-MoE-A2.7B, we ran the redistribution analysis at layers
0, 6, 12, 18, and 23, computing per-expert statistics over the
WikiText-2 test corpus ($n = 34{,}016$ token-level expert
activations per layer). Table~\ref{tab:qwen_redist} reports mean gap
norm across the 60 routed experts, Spearman $\rho$ between routing
weight and gap norm, and Spearman $\rho$ between gap norm and
relative compensation.

\begin{table}[h]
\centering
\caption{Qwen1.5-MoE redistribution analysis. Correlations computed across $N = 60$ routed experts at each layer. Significance markers: $^{*}p < 0.05$; $^{**}p < 0.01$; $^{***}p < 0.001$.}
\label{tab:qwen_redist}
\small
\begin{tabular}{lccccc}
\toprule
 & L0 & L6 & L12 & L18 & L23 \\
\midrule
Mean gap norm   & $0.011$ & $0.122$ & $0.090$ & $0.239$ & $0.440$ \\
$\rho$(rw, gap norm) & $+0.558^{***}$ & $+0.638^{***}$ & $+0.632^{***}$ & $+0.596^{***}$ & $+0.744^{***}$ \\
$\rho$(gap norm, rel.~comp.) & $+0.076$ & $+0.394^{**}$ & $+0.388^{**}$ & $+0.349^{**}$ & $+0.282^{*}$ \\
\bottomrule
\end{tabular}
\end{table}

The gap-norm-to-relative-compensation correlation is non-significant
at Layer~0 ($\rho = +0.076$, $p = 0.563$) but significant
($p < 0.05$) from Layer~6 onwards, in contrast to OLMoE where this
correlation strengthens only at Layer~15. The mean gap norm scales
$\approx 38\times$ from Layer~0 ($0.011$) to Layer~23 ($0.440$),
comparable to OLMoE's $41\times$ scaling.

\section{Architecture Verification Procedures}
\label{app:verification}

For each cross-architecture replication, we ran a four-test
verification suite before collecting per-token metric ablation and
routing weight control data:

\begin{enumerate}
    \item \textbf{Per-token CE matches HuggingFace reference loss.}
    For one corpus sample, mean per-token cross-entropy over
    non-padding positions is compared to HuggingFace's
    \texttt{labels=input\_ids} loss. Tolerance: $10^{-2}$ for
    half-precision runs, $10^{-4}$ for fp32.
    \item \textbf{No stale state after clearing ablation hooks.}
    For one position, baseline loss, ablated loss, and post-clear
    loss are computed; baseline and post-clear must agree to within
    $10^{-3}$.
    \item \textbf{Position diversity.} Across 100 positions in one
    sample, distinct losses must exceed 50\% of positions.
    \item \textbf{Position-specific ablation effect.} Ablating
    expert $e$ at a position where $e$ is active must produce
    larger loss change than at a position where $e$ is inactive.
\end{enumerate}

All three models passed all four verification tests. For
DeepSeek-V2-Lite verification was repeated independently at each of
the 5 tested layers; Qwen and OLMoE verifications were performed
once at the most-tested middle layer.

\section{Progressive Ablation Per-Layer Trajectories}
\label{app:progressive}

For the progressive ablation experiment summarised in
Section~\ref{sec:results}, Table~\ref{tab:progressive_mean} reports
mean cumulative $\Delta\mathcal{L}$ (in nats) on OLMoE-1B-7B-0924
for each tested layer at each removal depth $k$, computed over
$n = 500$ WikiText-2 token positions per (layer, $k$) cell.
Table~\ref{tab:progressive_tails} reports the corresponding median
and P95 values, which clarify that the mean increase at Layer~15 is
driven by a heavy upper tail rather than a uniform shift. Median
$\Delta\mathcal{L}$ at $k = 7$ remains below $+0.02$ nats at every
tested layer including Layer~15, while P95 reaches $+2.633$ nats at
Layer~15 versus $\leq +0.709$ elsewhere. The redundancy cliff at
Layer~15 is concentrated on a minority of positions where the
ensemble does not absorb removal.

\begin{table}[h]
\centering
\caption{Mean cumulative $\Delta\mathcal{L}$ (nats) under progressive
ablation of the $k$ highest-routing-weight active experts at each
layer in OLMoE-1B-7B-0924. $n = 500$ token positions per cell.}
\label{tab:progressive_mean}
\small
\begin{tabular}{ccccccc}
\toprule
$k$ & L0 & L1 & L7 & L8 & L9 & L15 \\
\midrule
1 & $+0.009$ & $+0.018$ & $+0.009$ & $+0.015$ & $+0.029$ & $+0.082$ \\
2 & $+0.013$ & $+0.031$ & $+0.016$ & $+0.028$ & $+0.060$ & $+0.155$ \\
3 & $+0.016$ & $+0.035$ & $+0.010$ & $+0.045$ & $+0.061$ & $+0.214$ \\
4 & $+0.019$ & $+0.051$ & $+0.022$ & $+0.054$ & $+0.073$ & $+0.290$ \\
5 & $+0.029$ & $+0.058$ & $+0.022$ & $+0.061$ & $+0.080$ & $+0.354$ \\
6 & $+0.021$ & $+0.058$ & $+0.025$ & $+0.062$ & $+0.078$ & $+0.403$ \\
7 & $+0.028$ & $+0.068$ & $+0.020$ & $+0.070$ & $+0.083$ & $+0.431$ \\
\bottomrule
\end{tabular}
\end{table}

\begin{table}[h]
\centering
\caption{Median and P95 cumulative $\Delta\mathcal{L}$ (nats) at
$k = 7$ for each tested layer. Median values near zero throughout
indicate the mean increase at Layer~15 is concentrated in an upper
tail rather than a uniform shift.}
\label{tab:progressive_tails}
\small
\begin{tabular}{lcccccc}
\toprule
 & L0 & L1 & L7 & L8 & L9 & L15 \\
\midrule
Median & $+0.000$ & $+0.003$ & $+0.001$ & $+0.008$ & $+0.012$ & $+0.018$ \\
P95    & $+0.433$ & $+0.398$ & $+0.473$ & $+0.641$ & $+0.709$ & $+2.633$ \\
\bottomrule
\end{tabular}
\end{table}

\end{document}